\documentclass[letterpaper, 10 pt, conference]{ieeeconf}  

\IEEEoverridecommandlockouts                              

\usepackage{xcolor}
\usepackage[linesnumbered,ruled,vlined]{algorithm2e}
\usepackage{graphicx}
\usepackage{multirow}
\usepackage[para,online,flushleft]{threeparttable}

\usepackage{amsmath}
\usepackage{color}
\usepackage{amssymb}
\usepackage{booktabs}
\usepackage[colorlinks, linkcolor=red]{hyperref}
\usepackage{array}
\usepackage{float}
\usepackage{marvosym}
\SetKwInput{KwInput}{Input}                
\SetKwInput{KwOutput}{Output} 

\newcolumntype{C}[1]{>{\centering\arraybackslash}p{#1}}

\title{\LARGE \bf
SKU-Patch: Towards Efficient Instance Segmentation\\ for Unseen Objects in Auto-Store
}

\author{
Biqi Yang\textsuperscript{*}, Weiliang Tang\textsuperscript{*}, Xiaojie Gao, Xianzhi Li, Yun-Hui Liu, Chi-Wing Fu and Pheng-Ann Heng
\thanks{
* Equal contributions to the work. This work was supported by InnoHK of the Government of Hong Kong via the Hong Kong Centre for Logistics Robotics. B. Yang, W. Tang, X. Gao, C.-W. Fu and P.-A. Heng are with the Department of Computer Science and Engineering, The Chinese University of Hong Kong. P.-A. Heng is also with Guangdong-Hong Kong-Macao Joint Laboratory of Human-Machine Intelligence-Synergy Systems, Shenzhen Institutes of
Advanced Technology, Chinese Academy of Sciences. X. Li is with the School of Computer Science and Technology, Huazhong University of Science and Technology. Y.-H. Liu is with the Department of Mechanical and Automation Engineering, The Chinese University of Hong Kong.} 
\thanks{}
}

\begin{document}
\maketitle
\thispagestyle{empty}
\pagestyle{empty}

\newcommand{\ourmethod}{SKU-Patch}

\begin{abstract}
In large-scale storehouses, precise instance masks are crucial for robotic bin picking but are challenging to obtain.
Existing instance segmentation methods typically rely on a tedious process of scene collection, mask annotation, and network fine-tuning for every single Stock Keeping
Unit (SKU).
%
This paper presents~\ourmethod, a new patch-guided instance segmentation solution, 
leveraging only a few image patches for each incoming new SKU to predict accurate and robust masks, without tedious manual effort and model re-training.
%
Technical-wise, we design a novel transformer-based network with (i) a patch-image correlation encoder to capture multi-level image features calibrated by patch information and (ii) a patch-aware transformer decoder with parallel task heads to generate instance masks.
Extensive experiments on four storehouse benchmarks manifest that~\ourmethod~is able to achieve the best performance over the state-of-the-art methods.
Also,~\ourmethod~yields an average of nearly 100\% grasping success rate on more than 50 unseen SKUs in a robot-aided auto-store logistic pipeline, showing its effectiveness and practicality.
\end{abstract}



\section{INTRODUCTION}
\label{sec:intro}
In Automatic Storehouse~(auto-store), Stock Keeping Unit (SKU) refers to a unique identifier of each product in the inventory management.
To enable the robotic system to handle a large variety of SKUs, we need instance segmentation, a fundamental vision task, to segment each SKU in the camera view. 
%
%
By then, the instance-level masks can enable the robot arm to effectively pick the SKUs one by one.

\begin{figure}[t]
	\centering
 \includegraphics[width=1.0\linewidth]{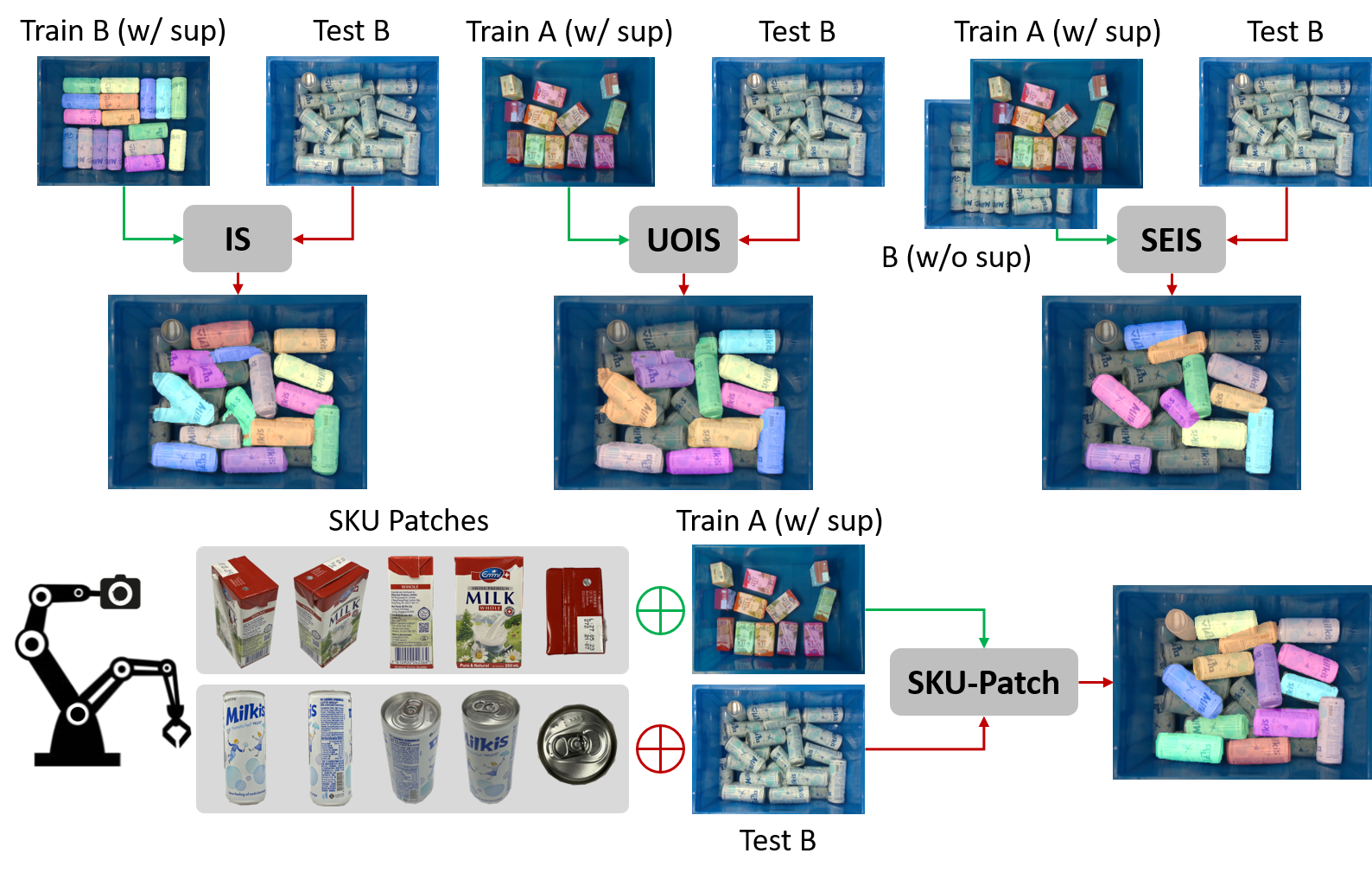}
 
	\vspace*{-3mm}
	\caption{Results on unseen SKUs produced by (i) the general Instance Segmentation (IS) method Mask R-CNN~\cite{he2017mask}, (ii) the Unseen Object Instance Segmentation (UOIS) method UOIS-2D~\cite{xie2021unseen}, (iii) the Self-Ensembling (SEIS) method SESR~\cite{yang2022sesr}, and 
 (iv)~\ourmethod~(ours). Clearly,~\ourmethod\ can generate more precise masks by leveraging only a few easily-obtainable SKU patches. 
 ``w/ sup'' and ``w/o sup'' mean with and without supervision.
	}
	\vspace*{-6mm}
	\label{fig: teaser}
\end{figure}

Recent deep-learning approaches show promising performance for general Instance Segmentation (IS, Fig.~\ref{fig: teaser} top-left)~\cite{he2017mask,liu2018path,chen2018masklab}.
Yet, it remains challenging to tackle Unseen Object Instance Segmentation~(UOIS, Fig.~\ref{fig: teaser} top-middle), i.e., testing SKUs different from training SKUs.
While some recent works on UOIS~\cite{xie2020best,back2021unseen,xie2021unseen,zhang2022unseen} start to address the seen-unseen domain gap, they may not work well for large-scale storehouses with a huge amount and variety of SKUs.

To improve the segmentation accuracy on unseen SKUs, the Self-Ensembling Instance Segmentation approach~(SEIS, Fig.~\ref{fig: teaser} top-right)~\cite{gu2021class,yang2022sesr} is proposed.
By acquiring quite a number of images prepared for each unseen SKU, teacher-student pseudo labeling can be exploited to enable the pre-trained model to fit the unseen objects.
Though human annotations are not required, the segmentation performance is still not satisfactory for successful grasping in practice, especially for crowded and chaotic scenes.
%
Besides, existing solutions have two other issues for handling the challenging large-scale auto-store environment.
First, they heavily rely on manual efforts to tediously prepare a large volume of data 
for every single SKU.
Second, for every new incoming SKU, the model has to be fine-tuned; doing so unavoidably
lower the practicality, efficiency, and scalability.

To meet these challenges, we develop~\ourmethod, a new patch-guided instance segmentation (Fig.~\ref{fig: teaser} bottom) solution for auto-store.
%
Without requiring
scene-level data collection and tedious human annotations, we employ only a \textit{few image patches} prepared for each new SKU for predicting the object masks.
%
Compared with IS,~\ourmethod\ can well generalize to efficiently handle a large variety of unseen SKUs with much lower data-collection requirements.
Comparing with UOIS,~\ourmethod\ harnesses the strength of instance patches and is able to produce convincing masks accurately for auto-store picking. Further, using a well-trained~\ourmethod\ network for a known SKU database, we can directly deploy it to the robot system for real-time usage, without requiring any re-training and parameter tuning as in SEIS.

As shown in Fig.~\ref{fig: teaser}, the SKU patch provides informative color and texture information of the target object instance. 
Intuitively, these features are strong prior knowledge that can guide the segmentor to achieve annotation-free and precise predictions on the unseen SKUs.
Importantly, SKU patches can be collected with very little effort without requiring any image annotation: it takes only a few seconds to capture 10 patches for an SKU with an industrial collection system~\cite{wei2019rpc,liu2020keypose}. 
%
So, the challenges lie in how to extract vision contexts from the SKU patches and use the patch-level information to guide the image-level instance segmentation.
Technical-wise, we propose a patch-guided transformer framework designed for SKU patches.
Our network consists of a patch-image correlation encoder, a patch-aware image decoder, and task heads. The encoder learns multi-level patch-image correlation through the cross-attention mechanism and outputs the calibrated image feature pyramid. The decoder aggregates the image features and learns a fixed number of compact instance embeddings. Then, the classification, box, and mask heads make predictions for each embedding. 

%
Extensive experiments show that \ourmethod\ is an effective solution for auto-store.
It can robustly provide instance masks for unseen SKUs, using only a few image patches.
Also, our Transformer~\ourmethod\ network achieves top precision and recalls on four benchmarks, outperforming
the state-of-the-art approaches. 
To test its practical performance in an auto-store setting, we deploy it on a real robot arm and use it to grasp more than 50 collected unseen SKUs.
SKU-Patch has a success rate of nearly 100\% on average for these wide varieties of objects of different appearances.

Our contributions can be summarized as follows:
\begin{itemize}
    \item We formulate~\ourmethod, a new solution for segmenting mass unseen SKUs in auto-store, requiring only a few image patches and relieving us from tedious pixel-level annotations and incessant network tuning.
    \item We design the novel transformer-based~\ourmethod\ network to correlate local instance patches and global cluttered scenes, so as to fully exploit the patch information for accurate mask predictions of unseen objects.
    \item Extensive experiments on four benchmark datasets  show the top performance of~\ourmethod, compared with recent SOTA methods.
    Robotic bin picking demos on more than 50 different unseen SKUs further demonstrate the strong capability of~\ourmethod.
\end{itemize}

\section{Related Works}
\label{sec:rw}

\subsection{Instance Segmentation for Auto-Store}
In storehouse environments, instance segmentation is crucial for supporting robotic grasping and suction~\cite{hasegawa2019graspfusion,li2021simultaneous,fang2018multi,xu2021pois}. The challenges lie in two aspects. 
The first is on handling cluttered scenarios with severe occlusion.
\cite{gupta2014learning, wada2019joint} enable deep neural networks to leverage both RGB and depth information.
\cite{wada2018instance,back2021unseen} harness the amodal concepts to improve segmentation performance on occluded regions.
\cite{neven2019instance, ito2020point} directly predict instance masks in one stage.
The second is on adapting to unseen objects with least data requirement. Some methods use sim-to-real technique~\cite{danielczuk2019segmenting,xiang2020learning,xie2020best,back2020segmenting,xie2021unseen,back2021unseen,li2022sim} while others~\cite{yang2022sesr,liu2022unseen} harness the strength of knowledge distillation, trained in a weakly-supervised manner. 
%
Meanwhile, various datasets are released for facilitating studies on instance segmentation for auto-store. \cite{follmann2018mvtec, wei2019rpc, goldman2019precise} provide large-scale annotated images with daily products. Xie et al.~\cite{xie2021unseen} builds a large-scale synthetic table-top dataset, which can be used for network pretraining.

\subsection{Few-shot Instance Segmentation}
Few-shot instance segmentation aims to leverage a few labeled samples (support images) on different parts of the network to achieve segmentation on unseen-class images. Some works~\cite{yan2019meta,michaelis2018one} on the prediction heads, some~\cite{fan2020fgn, fan2020few, zhang2020few} on the regional proposal network, while others~\cite{nguyen2021fapis, han2023reference, li2020one, chen2021dual, han2022few} focus on the backbone. Incorporation mechanisms mainly include simple feature concatenation~\cite{fan2020few, yan2019meta, xiao2020few}, prototypical-based feature aggregation~\cite{ganea2021incremental, wu2021universal}, and using attention network~\cite{han2022meta, han2022few, chen2021adaptive, chen2021dual, doersch2020crosstransformers, hsieh2019one}. 
Despite their progress, the above methods still require network fine-tuning at test time (which is not efficient in the auto-store setting) to handle each unseen class. In contrast, we design~\ourmethod~to take only a few image patches of the unseen SKU to predict the instance masks. 

\section{Method}
\label{sec:method}

\subsection{Overview}
We propose \ourmethod, a new patch-guided instance segmentation approach for unseen SKU instance segmentation.
Besides scene images, we take only a few additional SKU patches as input. 
 It learns to calibrate the image feature under the guidance of SKU patches. 
Based on the calibrated features, it can well segment the SKU instances corresponding to the given SKU patches.
Once the \ourmethod~model is trained, it can be directly applied to arbitrary unseen SKUs, without tedious data collection, label annotation, and model fine-tuning for each upcoming new SKU.
%
%
%
%
%
%
%

\begin{figure}[t]
	\centering
	\includegraphics[width=1.0\linewidth]{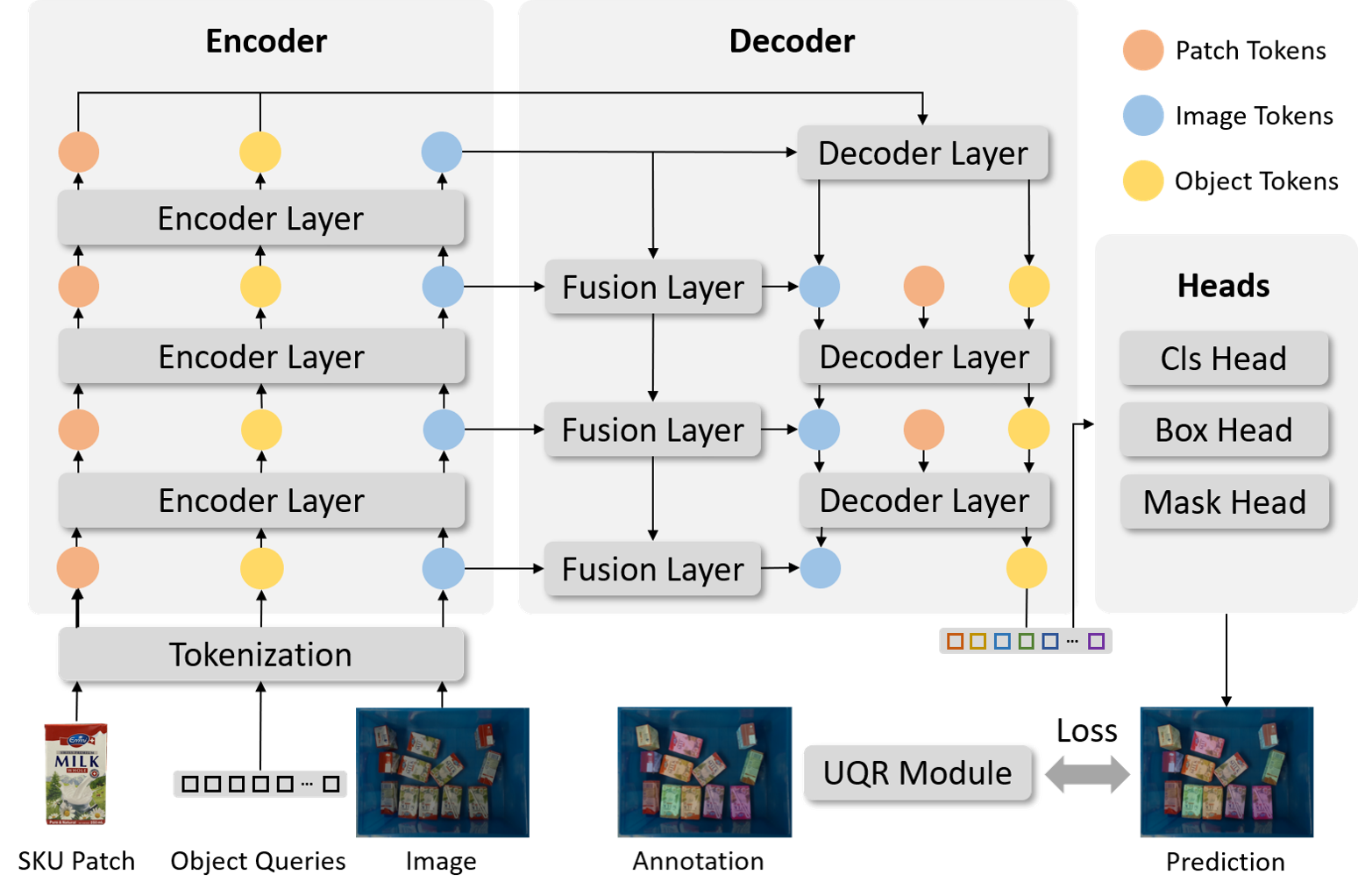}
	\vspace*{-8mm}
	\caption{The pipeline of our proposed network, which is built with a patch-image correlation encoder, a patch-aware decoder, and three task heads.
	}
	\vspace*{-6mm}
	\label{fig: pipeline}
\end{figure} 

The main challenge is how to extract patch features to calibrate the image features. To tackle this correlation-learning problem, we design a patch-based transformer architecture, refer to Fig.~\ref{fig: pipeline}. 
Similar to~\cite{carion2020end,zhu2020deformable,gao2021fast}, we tokenize the input image and an SKU patch for feature extraction, and use learnable object queries to predict the instance-level categories, positions, and masks. 
Let $z_I\in\mathbb{R}^{d\times N_I}$ (blue), $z_P\in\mathbb{R}^{d\times N}$ (orange), and $z_O\in\mathbb{R}^{d\times K}$ (yellow) be the \textit{Image}, \textit{Patch}, and \textit{Object} Tokens, respectively, where $N_I$ and $N$ denote the sequence length of the image and patch tokens respectively, $d$ denotes the feature dimension, and $K$ denotes the number of object queries used for segmentation. 
In Fig.~\ref{fig: pipeline}, our network consists of three parts.
%
First is a patch-image correlation encoder (Sec.~\ref{subsec:encoder}). It takes $\{z_I,z_P,z_O\}$ as inputs and hierarchically relates the local patches with the global scene for token calibration. 
The encoder outputs the multi-scale image features calibrated by the patch tokens and the high-level object tokens. 
Second is the patch-aware image decoder (Sec.~\ref{subsec:decoder}), it aggregates the image features with a pyramid fusion block and stacks deformable attention decoding layers to extract specified instance embeddings in a coarse-to-fine manner. 
Last is a set of three parallel task heads (Sec.~\ref{subsec:heads}), for classification, box regression, and mask generation. 
%
We use the Unified Query Representation (UQR) module~\cite{dong2021solq,yu2022soit} to convert the 2D ground truth mask to a 1D vector, so that it can well align with the mask head output.

The above basic network takes one image and a single patch as input. In the auto-store environments, the number of patches per SKU, however, varies. In Sec.~\ref{subsec:Kshot}, we propose an $N$-to-1 module to combine $N$ patches into an aligned representation, allowing flexible handling of different patch numbers.

\subsection{Patch-Image Correlation Encoder}
\label{subsec:encoder}
The encoder is designed to maximally exploit the information of a patch to guide the image feature extraction. It contains a set of hierarchical basic layers, each taking $\{z_I^i,z_P^i,z_O^i\}$ as inputs to produce calibrated $\{z_I^{i+1},z_P^{i+1},z_O^{i+1}\}$, where $i$ denotes the layer index.

In each layer, we design two cross-attention modules. 
First is the \textit{Patch-Image} cross-attention to enhance both the patch and image features, i.e., to highlight the image regions highly similar to the given patch (Eq.~\ref{eq: p-i}) and to augment the patch tokens by features of in-scenario instances (Eq.~\ref{eq: i-o}). 
With $\mathbf{K}, \mathbf{Q}, \mathbf{V}$ as three parallel Feed Forward Networks (FFNs) to generate key, query, and value in attention mechanism, the procedure can be detailed as:
\begin{equation}
\label{eq: p-i}
{z_I^{i+1}}={\rm Softmax}(\frac{\mathbf{Q}({z_I^i})\mathbf{K}^\top({z_P^i})}{\sqrt{d}})\mathbf{V}({z_P^i})
\end{equation}
\begin{equation}
\label{eq: i-o}
{z_P^{i+1}}={\rm Softmax}(\frac{\mathbf{Q}({z_P^i})\mathbf{K}^\top({z_I^{i+1}})}{\sqrt{d}})\mathbf{V}({z_I^{i+1}}),
\end{equation}
where $d$ is the output dimension of FFNs.
We then use an intermediate self-attention to capture long-range information within image features $z_I^{i+1}$ with Swin-Transformer~\cite{liu2021swin}.  
Second is the \textit{Image-Object} cross-attention to convey the contents from $z_I^{i+1}$ to $z_O^{i+1}$. Each object token specifies an instance, thus we harness the cross-attention to channel-wise aggregate the image features for $K$ different object tokens:
\begin{equation}
\begin{aligned}
{z_O^{i+1}}={\rm Softmax}(\frac{\mathbf{Q}({z_O^i})\mathbf{K}^\top({z_I^{i+1}})}{\sqrt{d}})\mathbf{V}({z_I^{i+1}}).
\end{aligned}
\end{equation}
We then build a self-attention for $z_O^{i+1}$ to help distinguish and diversify the instance. 

In practice, we set the number of layers $L$ as 4 and repeatedly stack four patch-guided transformer layers to construct our encoder. It hierarchically extracts multi-scale $\{z_I^i\}_{i=1}^L$ and the highest-level object embedding $z_O=z_O^L$.

%

%

%


\subsection{Patch-aware Transformer Decoder}
\label{subsec:decoder}

The decoder is designed to fuse multi-scale $\{z_I^i\}_{i=1}^L$ output from the encoder and iteratively refine $z_O$ with patch awareness. Based on the standard DETR decoder~\cite{carion2020end}, we propose the following adaptations.

First, to leverage the multi-scale image contents, we utilize a pyramid feature fusion technique. 
In each fusion layer, we bilinearly interpolate the low-resolution $z_I^{i+1}$ with an upsampling factor of two, add it with the high-resolution $z_I^i$ pre-processed by the ResNet bottelneck~\cite{he2016deep}, and then pass it through another bottleneck layer to produce the fused $z_I^i$. Compared with DETR, which only utilizes a single-scale image feature at the highest level, we progressively aggregate multi-scale image information in a top-down manner.

The second design is the patch-aware decoder layer. 
In each layer, we build a cross-attention module to further enhance $z_I^i$ by the patch tokens $z_P^i$ of the same level:
\begin{equation}
\begin{aligned}
{z_I^i}={\rm Softmax}(\frac{\mathbf{Q}({z_I^i})\mathbf{K}^\top({z_P^i})}{\sqrt{d}})\mathbf{V}({z_P^i}).
\end{aligned}
\end{equation}
With the patch-aware image features, we then use the deformable attention~\cite{zhu2020deformable} to refine $z_O^i$. 
Compared to standard attention, which has to look over all possible spatial locations in the image space, we only attend to a small set of key sampling points, thereby greatly improving the overall efficiency and speeding up the convergence.
Specifically, deformable attention predicts and samples $D$ important \textit{Image} Tokens $\{z_{I,j}^i\}_{j=1}^{D}$ and update $z^i_O$ as follows:
\begin{equation}
{z^i_O}={\rm Softmax}(\frac{\mathbf{Q}(z^i_O)\mathbf{K}^\top(\{z_{I,j}^i\}_{j=1}^{D})}{\sqrt{d}})\mathbf{V}(\{z_{I,j}^i\}_{j=1}^{D})
\end{equation}

We iteratively pass the $z_I^i$ and $z_O^i$ through the stacked decoder layer, until we obtain the final $z_O=z_O^1$, representing $K$ specified instance-level embeddings.

\begin{table*}[t]
	\centering
	\caption{Comparing the instance segmentation performance of our method against others on unseen SKUs of auto-store dataset.} 
	\vspace{-3mm}
	\label{table: AutoStore}
	\begin{threeparttable}
		\resizebox{0.8\textwidth}{!}{%
			\begin{tabular}{c||cc|cc|cc|cc}
				\hline
				\multirow{2}{*}{Method} & \multicolumn{2}{c|}{{mAP}$_{50}$} & \multicolumn{2}{c|}{{mAP}$_{75}$} & \multicolumn{2}{c|}{{mAP}$_{50:95}$} & \multicolumn{2}{c}{Recall} \\ \cline{2-9} 
    
				{} & easy & hard & easy & hard & easy & hard & easy & hard \\\hline
    
				Mask R-CNN (CVPR'17)~\cite{he2017mask} & 0.768 & 0.616 & 0.548 & 0.346 & 0.477 & 0.209 & 0.579 & 0.417 \\
    
				Mask R-CNN*\cite{he2017mask} & 0.807 & 0.695 & 0.645 & 0.532 & 0.520 & 0.418 & 0.630 & 0.527 \\
    
                UOIS-2D (TRO'21)\cite{xie2021unseen} & 0.826 & 0.710 & 0.629 & 0.537 & 0.538 & 0.465 & 0.637 & 0.558 \\

                UOIS-2D*\cite{xie2021unseen} & 0.913 & 0.794 & 0.725 & 0.680 & 0.632 & 0.619 & 0.710 & 0.681\\
                
                SESR (IROS'22)\cite{yang2022sesr} & 0.906 & 0.797 & 0.742 & 0.719 & 0.643 & 0.628 & 0.702 & 0.678 \\\hline
                
                Ours (1 patch) & \underline{0.946} & \underline{0.827} & \underline{0.841} & \underline{0.760} & \underline{0.730} & \underline{0.654} & \underline{0.821} & \underline{0.769} \\
                
				Ours (5 patches) & \textbf{0.953} & \textbf{0.833} & \textbf{0.862} & \textbf{0.794} & \textbf{0.778} & \textbf{0.689} & \textbf{0.842} & \textbf{0.800} \\\hline
			\end{tabular}
		}
	\end{threeparttable}
 \vspace{-3mm}
\end{table*}

\begin{figure*}[t]
	\centering
	\includegraphics[width=1.0\linewidth]{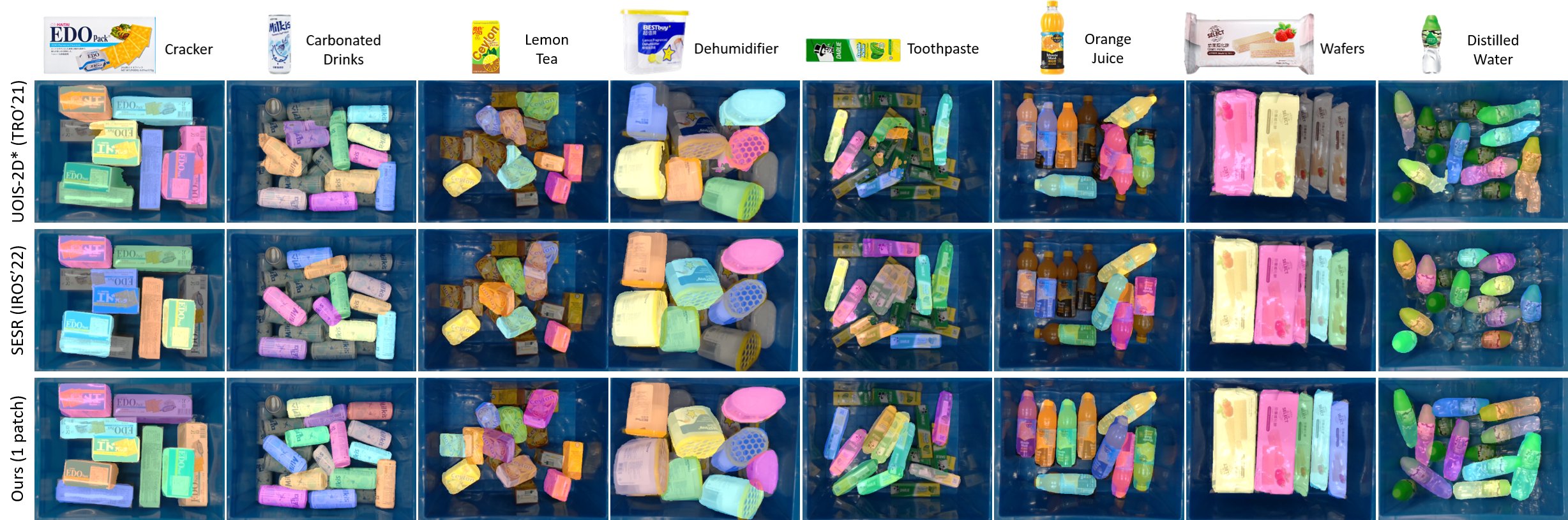}
	\vspace*{-7mm}
	\caption{Comparing instance segmentation results with fine-tuned UOIS method UOIS-2D*~\cite{xie2021unseen} and SEIS method SESR~\cite{yang2022sesr} on AutoStore dataset. Oue method gains convincing mask improvement on various unseen objects of different shape, size and texture.}
	\vspace*{-6mm}
	\label{fig: AutoStore}
\end{figure*} 

\subsection{Task Heads}
\label{subsec:heads}
Given $z_O\in\mathbb{R}^{d\times K}$, we feed each instance vector to three parallel task heads, each with four layers of FFNs, to predict a category result, a bounding box, and a semantic mask embedding, respectively. 
However, the mask embedding cannot be directly converted to a 2D result pixel level. Inspired by~\cite{fang2021you,song2021vidt}, we use the Unified Query Representation (UQR) module, which transforms the ground-truth mask to a vector, so that it can well align with our predicted embedding. 
Specifically, given the ground-truth mask $S$, its vector representation $F$ is obtained by sampling the low-frequency components $F=ASA^\top$, where $A$ is a matrix to apply discrete cosine transformation~\cite{ahmed1974discrete}. 
In this way, our predicted mask vectors from the FFN mask head can be supervised as 1D embeddings. 
Also, a mask vector can be recovered to a standard 2D map $\hat{S}$ through $\hat{S}=A^{-1}\hat{F}(A^\top)^{-1}$, where $\hat{F}$ is obtained by the inversely sampled $F$.

Given $K$ outputs and the corresponding annotations, we assign ground truths to the network predictions by bipartite matching with the Hungarian algorithm~\cite{carion2020end} to one-to-one calculate losses. 
To supervise the classifier, we use the Cross Entropy loss. For box supervision, we use the Smooth L1 loss together with Generalized IoU loss~\cite{rezatofighi2019generalized}. For the mask vector supervision, we use Smooth L1 loss.
\subsection{$N$-to-1 for Arbitrary Patch Number}
\label{subsec:Kshot}
In practice, the method should naturally adapt arbitrary patch numbers, since different SKUs can be accompanied with patch quantity variance. 
For example, SKUs of irregular shapes or complex textures may need sightly more patches for better description.
To extend the above method for arbitrary patch numbers, we design the $N$-to-1 module to generate the single representative patch feature $\hat{z}_P$ from the given $N \{z_P^j\}_{j=1}^N$.
Specifically, we first take the first patch $z_P^1$ as $\hat{z}_P$, which corresponds to the scenario when $N=1$. 
For $N>1$, we then repeatedly update $\hat{z}_P$ using the remaining $N-1$ patches as
\begin{equation}
\hat{z}_P={\rm Softmax}(\frac{\mathbf{Q}({\hat{z}_P})\mathbf{K}^\top({z_P^j})}{\sqrt{d}})\mathbf{V}({z_P^j}),j=2,...,N.
\end{equation}
The proposed $N$-to-1 module effectively integrates features of multiple patches and coherently generates a single patch representation regardless of the patch number $N$, thus can align SKU patches of different views with the lowest computation cost, leading to a scalable and efficient solution.

\section{Experiments and Results}
\label{sec:experiment}

\begin{table*}[t]
	\centering
	\caption{Comparing the detection performance of our method against others on unseen SKUs of the RPC dataset.}
	\vspace{-3mm}
	\label{table: RPC}
	\begin{threeparttable}
		\resizebox{0.98\textwidth}{!}{%
			\begin{tabular}{c||ccc|ccc|ccc}
				\hline
				\multirow{2}{*}{Method} & \multicolumn{3}{c|}{Split-20} & \multicolumn{3}{c|}{Split-50} & \multicolumn{3}{c}{Split-100} \\ \cline{2-10} 
    
				{} & mAP50 & mAP75 & Recall & mAP50 & mAP75 & Recall & mAP50 & mAP75 & Recall \\\hline
    
				Rotated Faster R-CNN~\cite{liu2017rotated} & 0.614 & 0.508 & 0.526 & 0.519 & 0.397 & 0.418 & 0.306 & 0.114 & 0.135 \\
    
				DPSNet (MM'20)~\cite{zhang2020iterative} & 0.687 & 0.548 & 0.566 & 0.544 & 0.487 & 0.493 & 0.427 & 0.296 & 0.338 \\
    
                CLCNet (AAAI'21)~\cite{cai2021rethinking} & 0.708 & 0.590 & 0.594 & 0.569 & 0.495 & 0.498 & 0.413 & 0.275 & 0.300 \\\hline
                
                Ours (1 patch) & \underline{0.725} & \underline{0.628} & \underline{0.610} & \underline{0.607} & \underline{0.538} & \underline{0.552} & \underline{0.485} & \underline{0.376} & \underline{0.390} \\
                
				Ours (5 patches) & \textbf{0.731} & \textbf{0.635} & \textbf{0.623} & \textbf{0.621} & \textbf{0.549} & \textbf{0.578} & \textbf{0.492} & \textbf{0.395} & \textbf{0.393} \\\hline
			\end{tabular}
		}
	\end{threeparttable}
        \vspace{-5mm}
\end{table*}

\subsection{Implementation Details and Evaluation Metrics}
%
%
We implemented our framework with  PyTorch~\cite{paszke2019pytorch} and trained it for 50 epochs with the AdamW~\cite{loshchilov2017decoupled} optimizer using an initial learning rate of $10^{-4}$. For AutoStore dataset~\cite{yang2022sesr} with 1,000 images, the training takes about 17h and the inference runs at around 11.0 Fps on a single GeForce GTX 1080 Ti GPU. We use Swin-T (Tiny) as our base architecture of Swin Transformer and initialize the length of the object queries $K$ as 200, i.e., the maximum predicted instance number is 200.
%
To evaluate the quality of our predicted instance masks, we adopt precision and recall as the evaluation metrics. 
%
\begin{figure}[t]
	\centering
	\includegraphics[width=0.97\linewidth]{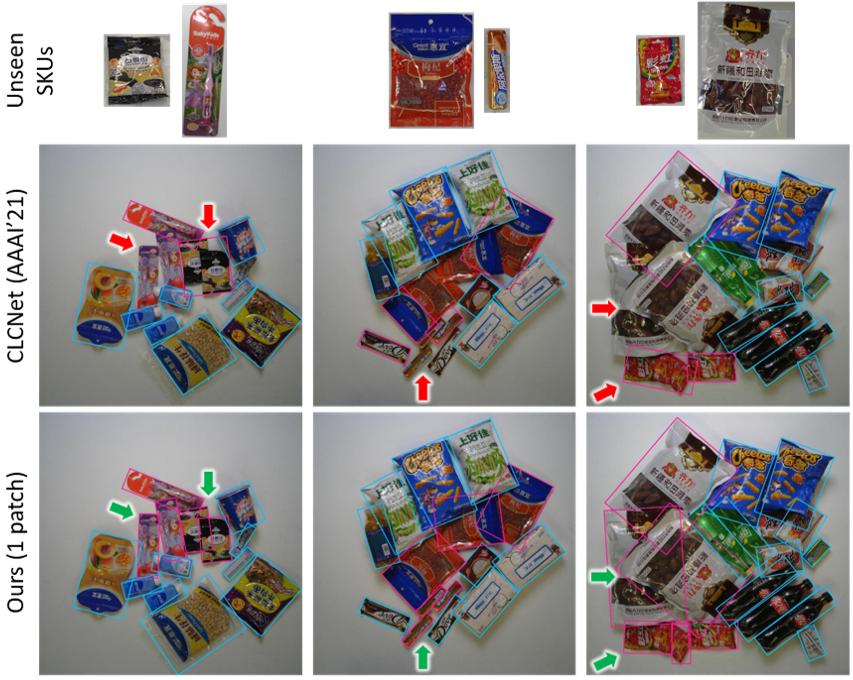}
	\vspace*{-3mm}
	\caption{Comparing results with CLCNet~\cite{cai2021rethinking} on typical RPC Split-20 scenes, where seen SKUs colored with blue and unseen SKUs colored with pink. Please follow the arrows for detailed comparison.}
	\vspace*{-2mm}
	\label{fig: RPC}
\end{figure} 
\begin{figure}[t]
	\centering
	\includegraphics[width=1.\linewidth]{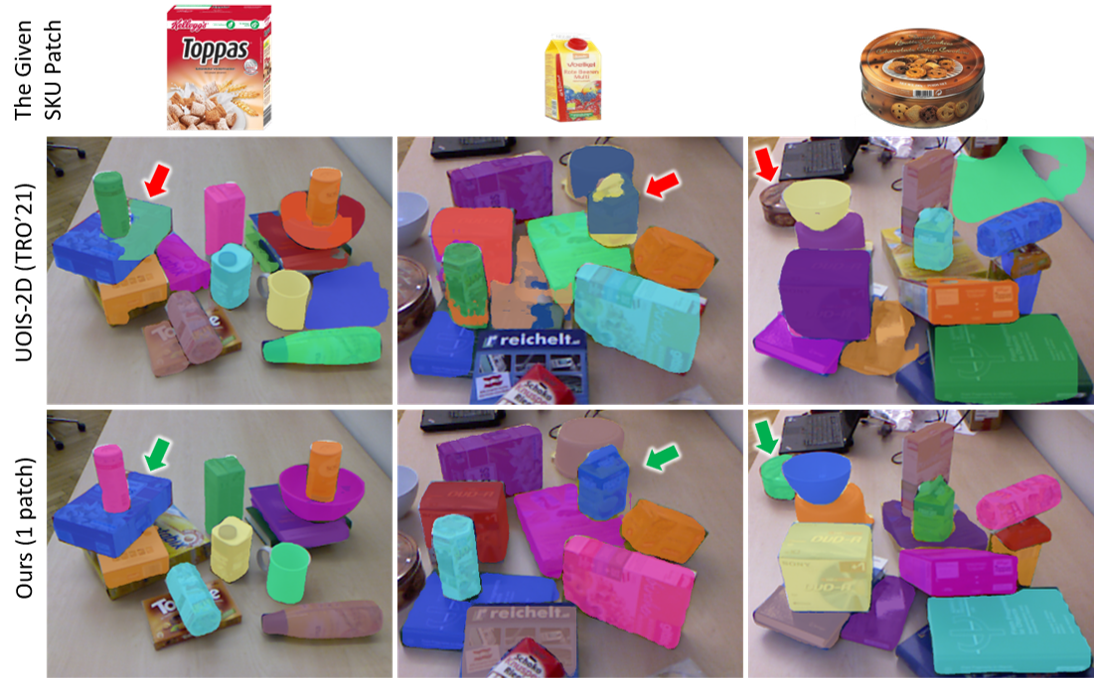}
	\vspace*{-7mm}
	\caption{Comparing results with UOIS-2D~\cite{xie2021unseen} on cluttered scenarios in OSD dataset, see the green arrows for effectiveness of patch guidance.}
	\vspace*{-6mm}
	\label{fig: OSD}
\end{figure} 
\begin{table}[t]
    \centering
    \caption{Comparing the UOIS performance of our method against others on OSD dataset with different input modalities.}
    \label{table: OSD}
    \vspace*{-3mm}
    \resizebox{0.45\textwidth}{!}{%
    \begin{tabular}{l||c||ccc}
        \hline
        {} & Modality & Precision & Recall & F-measure \\ \hline
        UOIS-2D (TRO'21)~\cite{xie2021unseen} & RGB & 0.723 & 0.656 & 0.670 \\ 
        UOIS-2D (TRO'21)~\cite{xie2021unseen} & RGBD & 0.807 & 0.805 & 0.799 \\
        UCN+ (CoRL'21)~\cite{xiang2020learning} & RGBD & 0.874 & 0.874 & 0.874 \\
        UOAIS (ICRA'22)~\cite{back2021unseen} & RGBD & 0.853 & 0.854 & 0.852\\ 
        FTEA+ (ICRA'23)~\cite{zhang2022unseen} & RGBD & \textbf{0.899} & \textbf{0.894} & \textbf{0.895}\\  \hline
        Ours+ (1 patch) & RGB & 0.862 & 0.840 & 0.851\\ \hline
    \end{tabular}
    }
    \vspace*{-6mm}
\end{table}
\subsection{Evaluation on AutoStore Dataset}
\label{exp: AutoStore}
The AutoStore dataset contains 50 kinds of daily supermarket SKU in total. There are 1,000 images, 20 per SKU, half for easy cases and half for hard cases, according to the clutter level. Each SKU is accompanied by five representative patches, which are captured by a camera fixed on the robot arm from different views. 

To demonstrate our method can effectively segment unseen SKUs, we split the dataset into five folds, each with 40 training categories and the rest 10 for inference. For each method, we conduct five experiments and calculate the mean performance in Tab.~\ref{table: AutoStore}. 
We compare our method with the general Instance Segmentation (IS) network Mask R-CNN~\cite{he2017mask}, the Unseen Object Instance Segmentation (UOIS) network UOIS-2D~\cite{xie2021unseen}, and the recent Self-Ensembling (SEIS) network SESR~\cite{yang2022sesr}. Note that SESR has full access to the testing images but without annotations.
Also, we compare with Mask R-CNN* and UOIS-2D*, where * indicates that for each unseen SKU, we add five extra images with full mask supervision for fine-tuning, which can significantly improve the original performance on those SKUs. 
Our method has no access to any unseen SKU when training. We assume that a new coming SKU has no chance for image capturing and annotation collection, only patch information can be provided.

In Tab.~\ref{table: AutoStore}, our method achieves the best result in both precision and recall. Especially for the hard cases, where objects are cluttered or stacked with occlusion, our method can still provide convincing masks. Refer to Fig.~\ref{fig: AutoStore} for those challenging scenarios. Comparing with UOIS-2D* (w/ fine-tuning) and SESR (can access un-annotated images of unseen SKUs when training), ours can well adapt to a large range of SKUs of different sizes, shapes, and textures.
\subsection{Evaluation on RPC Dataset}
\label{exp: RPC}
The Retail Product Checkout (RPC) dataset~\cite{wei2019rpc} includes 200 retail SKUs. In total, there are 30,000 images containing 367,935 instances. 
%
For each SKU, the RPC dataset provides abundant exemplar patches of different object poses. To adapt our method, we choose five basis patches for each SKU, more details can be found in the supplemented video.
We split the RPC dataset into training and inference sets. We create Split-20, -50 and -100, where the numbers count unseen categories over the total 200 categories. Noted that the training images only include instances of seen classes, an image containing both seen and unseen SKUs is assigned to the inference set. 
Statistically, the numbers of training images are 15333, 5663, and 1017 for Split-20, -50, and -100, where Split-100 is an extremely hard mode.


%
Since RPC only has box annotations, we omit the segmentation part in Sec.~\ref{subsec:heads} to modify the proposed method for object detection. Note that for all methods, we leverage rotated boxes for compact and tight bounding.
Apart from Rotated Faster R-CNN~\cite{liu2017rotated}, we also compare with two typical methods for auto-store object detection. DPSNet~\cite{zhang2020iterative} leverages iterative knowledge distillation to tackle distribution gap. CLCNet~\cite{cai2021rethinking} is a multi-task framework for simultaneous object detection and goods counting.

Tab.~\ref{table: RPC} shows the quantitative results. 
Comparing with others, the performance of our method downgrades less when the scenarios get harder (from Split-20 to -100), showing the robustness of \ourmethod. 
Fig.~\ref{fig: RPC} shows some typical results of Split-20, given merely one patch per unseen SKU, our method still achieves significant detection improvement (see the pink boxes), especially for objects closely-packed together with occlusion and appearance ambiguity, an SKU patch plays an critical role to separate instances apart.
\begin{table}[t]
    \centering
    \caption{Comparing the detection performance of our method against others on seen SKUs of SKU110K dataset.}
    \label{table: SKU110K}
    \vspace*{-3mm}
    \resizebox{0.5\textwidth}{!}{%
    \begin{tabular}{l||cccc}
        \hline
        {} & {mAP}$_{50}$ & {mAP}$_{75}$ & {mAP}$_{50:95}$ & Recall \\ \hline
        LTM (TPAMI'21)~\cite{9321141} & 0.578 & 0.509 & 0.453 & 0.492 \\ 
        HNAA (WACV'22)~\cite{cho2022densely} & 0.589 & 0.512 & 0.459 & 0.497\\ 
        BlackBox (RAL'22)~\cite{miller2022s} & 0.583 & 0.526 & 0.468 & 0.508\\ \hline
        Ours (1 patch) & \underline{0.592} & \underline{0.547} & \underline{0.480} & \underline{0.542}\\ 
        Ours (5 patches) & \textbf{0.598} & \textbf{0.550} & \textbf{0.485} & \textbf{0.545} \\ \hline
    \end{tabular}
    }
    \vspace*{-2mm}
\end{table}
\begin{figure*}[t]
	\centering
	\includegraphics[width=0.95\linewidth]{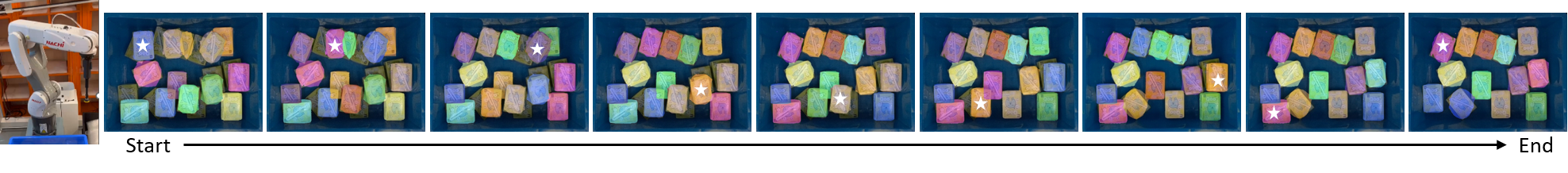}
	\vspace*{-5mm}
	\caption{A picking sequence according to the mask results generated by our method, where the object with a white star is the one to be picked.}
	\vspace*{-5.5mm}
	\label{fig: picking}
\end{figure*} 
\subsection{Evaluation on Cross (TOD$\rightarrow$OCID) Dataset}
\label{exp: OSD}
We further perform a cross-dataset experiment to show that \ourmethod\ helps overcome the sim-to-real data domain gap with patch guidance. In Sec.~\ref{exp: AutoStore}, our training and inference data are captured in the same environment whereas here, we train the model on the TOD dataset~\cite{xie2020best} with only synthetic images and test on 111 real images in the OSD dataset~\cite{richtsfeld2012segmentation}. Following the metrics in ~\cite{xie2020best,back2021unseen}, we quantitatively compare with UOIS methods in Tab.~\ref{table: OSD}, where + means zoom-in mask refinement. Worth noting that while most UOIS methods use RGBD images, ours use only single RGB modality with patch information. Refer to Fig.~\ref{fig: OSD}, where the arrows point the refined masks by patch guidance.
\subsection{Evaluation on SKU110K Dataset}
\label{exp: SKU}
Next, we show that \ourmethod not only works well for unseen SKUs but also gains improvement for seen objects with patch guidance. SKU100K is an object detection dataset~\cite{goldman2019precise} with 11,762 images and more than 1.7 million annotated bounding boxes. We randomly choose 427 images for training, 126 hard cases for inference, and manually crop 5 SKU patches per image.
We compare with three SOTA detection methods~\cite{9321141,cho2022densely,miller2022s} specifically designed for densely-packed scenes. Considering that the patch size is too small in the scenes, we use the crop and zoom-in pre-processing for better patch-image alignment. 
In Tab.~\ref{table: SKU110K}, our \ourmethod\ achieves the best result. Worth noting that SKU110K has an average of 148 instances per image, the extremely dense scenarios challenge the recall value, while our method significantly outperforms others on this metric.

\subsection{Ablation Studies}
\paragraph{Designs for Decoder}
Tab.~\ref{table: decoder} ablates our designs on auto-store dataset hard split with only one patch per unseen SKU.
We propose a pyramid feature fusion technique to aggregate multi-scale image features (Fuse). 
Furthermore, to decode a more informative image feature, we build a cross-attention module to calibrate the fused image features with patch guidance (Cross-A). 
Finally, we use the deformable attention~\cite{zhu2020deformable} instead of traditional full attention to construct each decoder layer (Deformable-A).

\begin{table}[t]
    \centering
    \caption{Ablation study of our designs for the decoder on auto-store dataset.}
    \label{table: decoder}
    \vspace*{-3mm}
    \resizebox{0.5\textwidth}{!}{%
    \begin{tabular}{c||ccc||ccc}
        \hline
         & Fuse & Cross-A & Deformable-A & mAP$_{50}$ & mAP$_{75}$ & Recall \\ \hline
        1 & - & - & - & 0.782 & 0.705 & 0.718\\ 
        2 & $\checkmark$ & - & - & 0.797 & 0.724 & 0.740  \\ 
        3 & $\checkmark$ & $\checkmark$ & - & 0.809 & 0.733 & 0.742 \\
        4 & $\checkmark$ & $\checkmark$ & $\checkmark$ & \textbf{0.827} & \textbf{0.760} & \textbf{0.769} \\ \hline
    \end{tabular}
    }\vspace*{-6mm}
\end{table}

\paragraph{$N$-to-1 Module}
We compare of our attention-based $N$-to-1 module with other feature combination methods, including feature adding (F-A) and feature momentum~\cite{yu2022graphfm} (F-M) in Tab.~\ref{table: Kto1}, meanwhile see effects of more patches. Since most auto-store goods have common and simple appearances, five patches can be enough and the improvement of more views for subtle object description is marginal.

\begin{table}[t]
    \centering
	\caption{Comparing different methods to process $N$ patches per unseen SKU on auto-store dataset.}
    \label{table: Kto1}
    \vspace*{-3mm}
	\begin{threeparttable}
		\resizebox{0.5\textwidth}{!}{%
			\begin{tabular}{c||ccc|ccc}
			\hline
			\multirow{2}{*}{} & \multicolumn{3}{c|}{$N=5$} & \multicolumn{3}{c}{$N=10$}\\ \cline{2-7} 
    
			{} & mAP$_{50}$ & mAP$_{75}$ & Recall & mAP$_{50}$ & mAP$_{75}$ & Recall \\\hline
    
			F-A & 0.784 & 0.707 & 0.704 & 0.809 & 0.721 & 0.728 \\
    
			F-M & 0.787 & 0.720 & 0.735 & 0.815 & 0.736 & 0.754 \\
    
            Ours & \textbf{0.827} & \textbf{0.760} & \textbf{0.769} & \textbf{0.832} & \textbf{0.766} & \textbf{0.769} \\\hline
			\end{tabular}
		}
	\end{threeparttable}
    \vspace*{-2mm}
\end{table}
\begin{table}[t]
    \centering
	\caption{Comparing the detection results with different lengths of object queries on two detection datasets.} 
	\vspace*{-3mm}
	\label{table: Det}
	\begin{threeparttable}
		\resizebox{0.5\textwidth}{!}{%
			\begin{tabular}{c||ccc|ccc}
			\hline
			\multirow{2}{*}{} & \multicolumn{3}{c|}{RPC Dataset Split-50} & \multicolumn{3}{c}{SKU110K Dataset} \\ \cline{2-7} 
    
			{} & mAP$_{50}$ & mAP$_{75}$ & Recall & mAP$_{75}$ & mAP$_{50}$ & Recall \\\hline
    
			$K=100$ & 0.603 & 0.532 & 0.550 & 0.574 & 0.529 & 0.536 \\
    
			$K=200$ & 0.607 & 0.538 & 0.552 & 0.592 & 0.547 & 0.542 \\
    
                $K=300$ & 0.607 & 0.543 & 0.552 & 0.616 & 0.563 & 0.568 \\\hline
			\end{tabular}
		}
	\end{threeparttable}
    \vspace*{-4mm}
\end{table}

\paragraph{Query Number $K$}


%
The number of object queries (i.e., the length of $z_O$) matters. Tab.~\ref{table: Det} shows the results of 100, 200 and 300 queries. While 100 is enough for RPC~\cite{wei2019rpc}, for extremely dense-packed scenes in SKU110K~\cite{goldman2019precise}, more queries obviously leads to better results.

\begin{table*}[htb]
    \centering
    \caption{Statics of grasping success rate of various unseen SKUs.}
    \label{table: picking}
    \vspace*{-3mm}
    \resizebox{0.95\textwidth}{!}{%
    \begin{tabular}{ccccccc}
        \hline
        Lotte & Vita & Lays & CocaCola & Nissin & Vinda & Hershey \\
        Pocky & Bottle Tea & Chip & Can Sprite & Cup Noodle & Roll Paper & Choco Bar \\ \hline
        94/94 & 126/126 & 65/65 & 133/133 & 109/109 & 78/78 & 158/158 \\
        100\% & 100\% & 100\% & 100\% & 100\% & 100\% & 100\% \\ \hline
    \end{tabular}
    }\vspace*{-5mm}
\end{table*}

\subsection{Robotic Picking Demonstrations}
We deploy our method in a storehouse setting using the Nachi MZ07 robot arm. Our method is responsible for generating precise segmentation results for a given scenario with unseen SKUs, then the system selects one of the results by heuristic height and point cloud smoothness analysis. Fig.~\ref{fig: picking} shows a  one-by-one picking sequence on an unseen paper-pack beverage. Please refer to the supplemented video, which quantitatively counts the grasping tries and success rate on different SKUs, accompanied with auto-store demos.

\section{Conclusion}
In this work, we propose \ourmethod, a new patch-guided instance segmentation
 solution for auto-store unseen SKUs. No need for any scene collection, image annotation, and network fine-tuning, \ourmethod\ takes only easy-to-obtain object patches to achieve convincing segmentation results on arbitrary unseen SKUs. We propose a Transformer-based network to support \ourmethod. Extensive experiments comparing with SOTA methods on existing benchmarks verify the effectiveness of our method. Robotic demonstrations further show that the \ourmethod\ solution is applicable for real-world real-time auto-store picking pipeline.

\bibliographystyle{IEEEtran}
\bibliography{ref}
\end{document}